\documentclass[letterpaper, 10 pt, conference]{ieeeconf}  % Comment this line out if you need a4paper

\IEEEoverridecommandlockouts                              % This command is only needed if 
\usepackage{graphics} % for pdf, bitmapped graphics files
\usepackage{epsfig} % for postscript graphics files
\usepackage{amsmath} % assumes amsmath package installed
\usepackage{amssymb}  % assumes amsmath package installed
\usepackage{color}
\usepackage{bm}
\usepackage{subfig}

\usepackage{soul}
    
\DeclareMathAlphabet\mathbfcal{OMS}{cmsy}{b}{n}

\title{\LARGE \bf
Active Structure-from-Motion for 3D Straight Lines*
}

\author{Andr\'e Mateus, Omar Tahri, and Pedro Miraldo% <-this % stops a space
\thanks{*This work was partially supported by the Portuguese FCT grants {\tt PD/BD/135015/2017} (through the NETSys Doctoral Program) \& {\tt SFRH/BPD/111495/2015}, and ISR/LARSyS Strategic Funding by the FCT project  {\tt PEst-OE/EEI/LA0009/2013}.}% <-this % stops a space
\thanks{A. Mateus \& P. Miraldo are with Institute for Systems and Robotics (ISR/IST), LARSyS, Instituto Superior T\'ecnico, Univ. Lisboa, Portugal \newline
        {\tt\footnotesize \{andre.mateus, pedro.miraldo\}@tecnico.ulisboa.pt}.}%
\thanks{O. Tahri is with INSA Centre Val de Loire, Universit\'e d'Orl\'eans, PRISME EA 
4229, Bourges, France 
        {\tt\footnotesize omar.tahri@insa-cvl.fr}.}%
}

\begin{document}

\maketitle
\thispagestyle{empty}
\pagestyle{empty}

%%%%%%%%%%%%%%%%%%%%%%%%%%%%%%%%%%%%%%%%%%%%%%%%%%%%%%%%%%%%%%%%%%%%%%%%%%%%%%%%
\begin{abstract}

A reliable estimation of 3D parameters is a must for several applications like planning and control, in which is included Image-Based Visual Servoing. This control scheme depends directly on 3D parameters, e.g. depth of points, and/or depth and direction of 3D straight lines.
Recently, a framework for \emph{Active Structure-from-Motion} was proposed, addressing the former feature type. 
However, straight lines were not addressed.
These are 1D objects, which allow for more robust detection, and tracking. In this work, the problem of \emph{Active Structure-from-Motion} for 3D straight lines is addressed.
An explicit representation of these features is presented, and a change of variables is proposed.
The latter allows the dynamics of the line to respect the conditions for observability of the framework.
A control law is used with the purpose of keeping the control effort reasonable, while achieving a desired convergence rate.
The approach is validated first in simulation for a single line, and second using a real robot setup.
The latter set of experiments are conducted first for a single line, and then for three lines.  

\end{abstract}

%%%%%%%%%%%%%%%%%%%%%%%%%%%%%%%%%%%%%%%%%%%%%%%%%%%%%%%%%%%%%%%%%%%%%%%%%%%%%%%%
\section{INTRODUCTION}
\label{sec:intro}

There are many robotic applications for which the recovery of 3D information is required.
Among those is Image-Based Visual Servoing, which consists in using visual information (e.g. image points) to control a robot \cite{chaumette2006}.
This kind of control is highly dependent on a good estimation of 3D parameters, for instance the depth of points, and/or the depth and direction vector of lines \cite{andreff2002}.

A way to estimate 3D information is Structure-from-Motion, which consists in estimating 3D parameters from images taken by a moving camera \cite{koenderink1991,bartoli2005}.
It is a well studied problem in Computer Vision, with emphasis on points and lines.
Even though those works present solutions to the problem, they suffer from considering either big displacement between views or more than two views.
Thus, preventing their use in Visual Servoing, since it requires continuous estimation.
In order to cope with these issues, some authors adopt filtering strategies, by taking measurements from consecutive camera images and the displacement of the camera between each frame (assumed to be known) \cite{matthies1989,soatto1996,civera2008,civera2010}. 

A non-linear state estimation scheme for unknown parameters is presented in \cite{deluca2008}, and is applied to the estimation of image points' depth, and focal length of a camera.
The observer is proved to be stable, as long as, the \emph{persistency of excitation condition} is satisfied.
More approaches, based on non-linear state estimators, are presented in \cite{dixon2003,morbidi2010,corke2010,sassano2010,martinelli2012,dani2012}. A comparison of a filtering solution with a non-linear estimation framework is presented in \cite{grabe2015}.

Some authors addressed the control of a camera to optimize 3D estimation (\emph{Active Vision} \cite{aloimonos1988}).
In \cite{chaumette1996}, a method to recover the 3D structure of points, cylinders, straight lines, and spheres is presented.
It consists in using the Implicit Function Theorem, to obtain an explicit expression of the 3D information as a function of image measurements, their time derivative, and the camera's velocity.
The active component (control law) attempts to simultaneously keep the features static in the image, and perform a desired trajectory (depending on the feature).
This approach requires the features time derivative (on the image plane) to be known, which are not directly available, and thus require discrete approximation.
Besides, the control law used does not give much insight on how it may affect the estimation error.

The optimization of the camera's motion, in order to have a desired estimation error response, is addressed in \cite{spica2013}, where a framework for \emph{Active Structure-from-Motion} is presented.
This is a general framework, that can be used for a wide variety of visual features, and its convergence behavior is well defined.
The framework has been applied to several feature types, like points, cylinders, spheres \cite{spica2014}, planes from measured image moments \cite{spica2015,spica2015b}, and rotational invariants \cite{tahri2015,tahri2017brunovsky}.
A method for coupling this framework with a Visual Servoing control scheme is presented in \cite{spica2017}.

\emph{Active Vision} has been addressed for a large set of visual features, yet straight lines have not been explored. 
The main focus has been the estimation of a point's depth.
However, in practice, it is useful to use richer features.
In particular, straight lines which are 1D objects, and whose one-to-one association from the image to the world is easier to obtain. Besides, lines are also easier to detect in images (e.g. Hough Transform \cite{matas2000}), and allow for a more robust tracking \cite{rosten2005}.
To the best of our knowledge, active vision for lines has been previously addressed by two works.
The first is \cite{chaumette1996}, whose shortcomings have already been stated.
The second is \cite{spica2014}, where lines are only considered indirectly, in the estimation of a cylindrical target radius.
In fact, the estimation scheme that allows a full line recovery is not explored, since the authors resort to a formulation specific to cylinders.

This work studies the problem of \emph{Active Vision} for 3D straight lines. 
Previous works represented these features as the intersection of two planes \cite{chaumette1996}. 
That representation is implicit, and does not have a direct counterpart in the image plane.
Thus, it is usually used together with the $(\rho,\theta)$ representation in the image plane, which even though is minimal, is also periodical \cite{andreff2002}.
Furthermore, the coupling of this two representations leads to complex dynamics.
In this work lines are represented by \emph{binormalized Pl\"ucker coordinates} \cite{andreff2002}, which are presented in Sec.~\ref{sec:dynamics}, along with their dynamics. 
This coordinates are explicit (a set of coordinates defines a single line); define lines everywhere in space (except if they contain the camera optical center); and have a direct representation in the image plane (the moment vector is the normal of the line's interpretation plane)\footnote{Plane containing the line and the optical center of the camera.
}.

State estimation is addressed by the framework in \cite{spica2013}, which is presented in Sec.~\ref{sec:active}.
Since, the dynamics of the \emph{binormalized Pl\"ucker coordinates} do not respect the requirements for convergence, a change of variables is presented in Sec.~\ref{sec:sfm_lines}.
Followed by the dynamics in the new coordinates, and the non-linear observer for 3D straight lines structure.

The proposed observer is validated in simulation for a single line, whose setup and results are presented in Sec.~\ref{sec:simulation}. 
The control law used for active structure estimation is also presented. 
This was designed to keep the control effort relatively low (in terms of the norm of the velocity vector), while achieving a desired convergence rate.
Those results are replicated with a real robotic platform \cite{messias2014}, in Sec.~\ref{sec:real}.
In Sec.~\ref{sec:3lines}, an application of the observer for recovering the 3D structure of three lines is presented.
Finally, the conclusions are presented in Sec.~\ref{sec:conclusions}.

\section{BACKGROUND}
\label{sec:back}

This section presents the theoretical background and concepts required for this work.
First, the representation of 3D straight lines and their dynamics are presented, followed by the framework for \emph{Active Structure-from-Motion}.

\subsection{Dynamics of the 3D Straight Lines}
\label{sec:dynamics}

\emph{Pl\"ucker} coordinates are a common way to represent 3D lines \cite{pottman2001}, given that they can represent lines everywhere in the space (except if they intersect the camera's optical center). Besides, they have a direct representation in the image plane, since they contain the normal vector to the interpretation plane.
This coordinates are defined as
\begin{equation}
    \mathbfcal{L} \sim \begin{bmatrix} \mathbf{u} \\ \mathbf{m} \end{bmatrix}, \quad \text{with} \quad \mathbfcal{L} \subset \mathcal{P}^5
\end{equation}
where $\mathcal{P}^5$ is the five-dimensional projective space, $\mathbf{u} \in \mathbb{R}^3$, and $\mathbf{m} \in \mathbb{R}^3$ represent the line's direction and moment respectively, up to a scale factor, and satisfy
\begin{equation}
    \mathbf{u}^T\mathbf{m} = 0.
\end{equation}

By normalizing the direction vector, we obtain the \emph{Euclidean Pl\"ucker coordinates} \cite{andreff2002}
\begin{equation}
    \begin{cases}
        \mathbfcal{L} = \begin{bmatrix} \mathbf{d} \\ \mathbf{n}  \end{bmatrix}\\
        \mathbf{d}^T \mathbf{n} = 0\\
        \mathbf{d}^T\mathbf{d} = 1,
    \end{cases}
\end{equation}
% \begin{equation}
%         \mathbfcal{L} = \begin{bmatrix} \mathbf{d} \\ \mathbf{n}  \end{bmatrix} \,\,\,        \text{with} \quad \mathbf{d}^T \mathbf{n} = 0 \,\,\, \text{and}\,\,\,
%         \mathbf{d}^T\mathbf{d} = 1,
% \end{equation}
where $\mathbf{d} = \frac{\mathbf{u}}{||\mathbf{u}||}$, and $\mathbf{n} = \frac{\mathbf{m}}{|| \mathbf{u} ||}$.
This representation is explicit, since a set of coordinates represents a single line.

Keep in mind that $\mathbf{n}$ represents the normal vector to the interpretation plane, thus it is possible to represent the projection of the line in the image $\mathbf{h}$ \cite{hartley2003} as
\begin{equation}
    \mathbf{h} = \frac{\mathbf{n}}{||\mathbf{n}||}.
\end{equation}

Let $l = ||\mathbf{n}||$ be the depth of the line.
If we recall that for every point $\mathbf{p} \in \mathbb{R}^3$ in the line it holds
\begin{equation}
    \mathbf{n} = \mathbf{p} \times \mathbf{d},
    \label{eq:depth_moment}
\end{equation}
and since $||\mathbf{d}|| = 1$. Then $||\mathbf{n}|| = ||\mathbf{p}||sin(\phi)$, with $\phi$ being the angle between the point and the direction.
If the depth of the line is considered to be the distance of the point closest to the origin, then the norm of $\mathbf{n}$ will always be equal to the depth.
Finally, the \emph{binormalised Pl\"ucker coordinates} \cite{andreff2002} can be defined as
\begin{equation}
    \mathbfcal{L} = \begin{bmatrix} \mathbf{d}\\ l\mathbf{h} \end{bmatrix}.
    \label{eq:biline}
\end{equation}

Let $\mathbf{v}_c = \left[ \bm{\nu}_c,\bm{\omega}_c \right]^T \in \mathbb{R}^6$ be the camera velocity, with $\bm{\nu}_c \in \mathbb{R}^3$ and $\bm{\omega}_c \in \mathbb{R}^3$ being the linear and angular velocities respectively. 
Then the dynamics of \eqref{eq:biline} are
\begin{align}
    \dot{\mathbf{d}} = & \; \bm{\omega}_c \times \mathbf{d} \label{eq:ddyn} \\
    \dot{\mathbf{h}} = & \; \bm{\omega}_c \times \mathbf{h} - \frac{\bm{\nu}_c^T\mathbf{h}}{l}( \mathbf{d}\times \mathbf{h}) \label{eq:hdyn} \\
    \dot{l} = & \; \bm{\nu}_c^T(\mathbf{d} \times \mathbf{h}). \label{eq:depthdyn}\
\end{align}

Let us consider the coordinates of a straight line to be our state vector. Thus, \eqref{eq:ddyn}, \eqref{eq:hdyn}, and \eqref{eq:depthdyn} give us the state equations. From those the state estimation scheme can be designed, whose background is presented in the next section.

\subsection{Active Structure-From-Motion}
\label{sec:active}

This section presents the framework for \emph{Active Structure-from-Motion} \cite{spica2013}.
Let $\mathbf{x} = \left[ \mathbf{s}, \bm{\chi} \right]^T \in \mathbb{R}^{m+p}$ be the state vector, whose dynamics are given by
\begin{equation}
    \begin{cases}
        \dot{\mathbf{s}} = \mathbf{f}_m (\mathbf{s}, \bm{\omega}_c) + \bm{\Omega}^T(\mathbf{s},\bm{\nu}_c)\bm{\chi} \\
        \dot{\bm{\chi}} = \mathbf{f}_u(\mathbf{s},\bm{\chi},\mathbf{v}_c),
    \end{cases}
\end{equation}
where $\mathbf{s} \in \mathbb{R}^m$ is the vector of the measurable components of the state, and $\bm{\chi} \in \mathbb{R}^p$ the unknown components.
Besides, let $\hat{\mathbf{x}} = [\hat{\mathbf{s}},\hat{\bm{\chi}}]^T$ be the estimated state, and $\mathbf{e} = [\tilde{\mathbf{s}},\tilde{\bm{\chi}}]^T$ be the state estimation error, with  $\tilde{\mathbf{s}} = \mathbf{s} - \hat{\mathbf{s}}$ and $\tilde{\bm{\chi}} = \bm{\chi} - \hat{\bm{\chi}}$.
Then, to recover the unknown quantities, the following estimation scheme can be used 
\begin{equation}
    \label{eq:framework_SOTA}
    \begin{cases}
        \dot{\hat{\mathbf{s}}} = \mathbf{f}_m (\mathbf{s}, \bm{\omega}_c) + \bm{\Omega}^T(\mathbf{s},\bm{\nu}_c)\hat{\bm{\chi}} + \mathbf{H} \tilde{\mathbf{s}}  \\
        \dot{\hat{\bm{\chi}}} = \mathbf{f}_u(\mathbf{s},\hat{\bm{\chi}},\mathbf{v}_c) + \alpha \bm{\Omega}(\mathbf{s},\bm{\nu}_c)  \tilde{\mathbf{s}},
    \end{cases}
\end{equation}
with $\mathbf{H}$ positive definite, and $\alpha > 0$. 

According to the \emph{persistency of excitation condition} \cite{deluca2008}, convergence of the estimation error to zero is possible, iff the matrix $\bm{\Omega}\bm{\Omega}^T$ is full rank.
Then, it is possible to prove that, the convergence rate will depend on the norm of that matrix, and in particular, on its smallest eigenvalue $\sigma_1^2$.
This depends on the measurable components of the state and on the camera's linear velocity, which can be optimized to control the convergence rate.

Let $\mathbf{U}\bm{\Sigma}\mathbf{V} = \bm{\Omega}$ be the singular value decomposition of matrix $\bm{\Omega}$, where $\bm{\Sigma} = [\mathbf{S} \quad \mathbf{0}]$, with $\mathbf{S} = diag(\{\sigma_i\})$, $i = 1,...,k$. Then, $\mathbf{H} \in \mathbb{R}^{m \times m}$ may be chosen as  
\begin{equation}
    \mathbf{H} = \mathbf{V} \begin{bmatrix} \mathbf{D}_1 & \mathbf{0} \\  \mathbf{0} & \mathbf{D}_2 \end{bmatrix} \mathbf{V}^T,
\end{equation}
where $\mathbf{D}_1 \in \mathbb{R}^{p\times p}$ is a function of the singular values of $\bm{\Omega}$, and $\mathbf{D}_2 \in \mathbb{R}^{(m-p) \times (m-p)}$ is set to be the identity matrix.
Following \cite{spica2013}, the former is defined as $\mathbf{D}_1 = diag(\{c_i\}), c_i > 0$ with $c_i = 2\sqrt{\alpha}\sigma_i$, and $i = 1,...,k$.
This choice prevents oscillatory modes, thus trying to achieve a critically damped transient behavior.

In order to get a desired rate, one can either tune the gain $\alpha$ and/or act on the input to lead the eigenvalue to a desired value. 
In this work, we focus on the latter.
Let us compute the total time derivative of the eigenvalues of matrix $\bm{\Omega\Omega}^T$.
Using the results in \cite{spica2013}, we conclude that 
\begin{equation}
   \dot{(\sigma_i^2)} = \mathbf{J}_{\bm{\nu}_c,i} \dot{\bm{\nu}_c} + \mathbf{J}_{\mathbf{s},i} \dot{\mathbf{s}} \quad \text{with} \; i = 1,..,k,
   \label{eq:dsigma}
\end{equation}
where matrices $\mathbf{J}_{\bm{\nu}_c,i}$  $\mathbf{J}_{\mathbf{s},i}$ are the Jacobian matrices yielding the relationship between the eigenvalues of $\bm{\Omega}\bm{\Omega}^T$, and the linear velocity and the measurable components of the state respectively.
In particular
\begin{equation}
    \mathbf{J}_{\bm{\nu}_c,i} =  \left[  \mathbf{v}_i^T \frac{\partial \bm{\Omega}\bm{\Omega}^T}{\partial \nu_x}\mathbf{v}_i, \mathbf{v}_i^T \frac{\partial \bm{\Omega}\bm{\Omega}^T}{\partial \nu_y}\mathbf{v}_i, \mathbf{v}_i^T \frac{\partial \bm{\Omega}\bm{\Omega}^T}{\partial \nu_z}\mathbf{v}_i  \right],
\end{equation}
with $\mathbf{v}_i$ being the normalized eigenvector associated with the $i^{th}$ eigenvalue of $\bm{\Omega}\bm{\Omega}^T$.
A differential inversion technique can be used to regulate the eigenvalues, by acting on vector $\dot{\bm{\nu}_c}$.
Notice that, when applying an inversion technique, it may not be possible to compensate for the second term on the right-hand side of \eqref{eq:dsigma}.
A way to compensate, for the effect of $\dot{\mathbf{s}}$ in the dynamics, is to enforce $\dot{\mathbf{s}} \simeq \mathbf{0}$.
In this work, that effect is compensated making use of the angular velocity as described in Sec.~\ref{sec:simulation}.

\section{ACTIVE STRUCTURE-FROM-MOTION USING 3D STRAIGHT LINES}
\label{sec:sfm_lines}

Sec.~\ref{sec:active} presents the framework for \emph{Active Structure-from-Motion}.
This consists of a full state observer, whose state is partially measured.
In this work, the measurement consists of vector $\mathbf{h}$, which is the normal vector to the interpretation plane of the line.
Besides $\mathbf{h}$, our representation consists of the direction vector and the line's depth.
These will only be retrieved from the estimated state.
A requirement is that the unknown variables appear linearly on the dynamics of the measurable ones.
From \eqref{eq:hdyn}, we can conclude that this does not happen.
In order to cope with this issue, the change of variables
\begin{equation}
    \bm{\chi} = \frac{\mathbf{d}}{l},
\end{equation}
can be considered.
Using \eqref{eq:depthdyn} and \eqref{eq:ddyn}, it is easy to conclude that
\begin{equation}
    \dot{\bm{\chi}} = \bm{\omega}_c \times \bm{\chi} - \bm{\chi} \bm{\nu}_c^T(\bm{\chi} \times \mathbf{h}).
    \label{eq:chidyn}
\end{equation}
Then, from \eqref{eq:framework_SOTA} and \eqref{eq:hdyn}, one can define
\begin{equation}
\label{eq:Omega_first}
\bm{\Omega} = - \bm{\nu}^T \mathbf{h} [\mathbf{h}]_\times,
\end{equation}
where $[\mathbf{h}]_\times$ is a $3\times3$ skew-symmetric matrix that linearizes the cross product ($\bm{\chi} \times \mathbf{h} = - [\mathbf{h}]_\times\bm{\chi}$). 

Convergence of the estimation error $\mathbf{e}$ to $0$ is possible \emph{iff} the square matrix $\bm{\Omega}\bm{\Omega}^T$ is full rank.
From \eqref{eq:Omega_first}, this is not the case, since its rank is equal to $2$ (this happens because of the multiplication by the skew-symetric matrix, which has, by definition, rank $2$).
In order to deal with this issue, the orthogonality of the \emph{Pl\"ucker} coordinates is explored. 
The obtained system belongs then to the large family of Differential Algebraic Systems \cite{campbell1982singular}. 
Since, the algebraic condition is  linear, a solution consists in expressing the coordinate of $\bm{\chi}$, corresponding to the coordinate of $\mathbf{h}$ with highest absolute value, as a function of the others.
This will not require a switching strategy\footnote{A switching strategy (in this context) is a change of the state variables in runtime, i.e., if the moment vector is not constant the coordinate with the highest absolute value is not always the same, and thus it  would be necessary to fix a different coordinate of $\bm{\chi}$.}, since the condition $\dot{\mathbf{h}} \simeq \mathbf{0}$ is enforce.
Thus, the coordinate with the highest absolute value is always the same throughout the task.
Notice that, by enforcing $\dot{\mathbf{h}} \simeq \mathbf{0}$, the angular velocity is used to maintain the interpretation plane of the line constant.
This allows us to compensate for the effects of $\dot{\mathbf{h}}$ in the dynamics of $\sigma_i^2$, and thus the linear velocity of the camera can be used to obtain a desired convergence rate.

\begin{figure*}
    %\vspace{-0.3cm}
    \centering
    \subfloat[Real and estimated state evolution.]{
        \includegraphics[width=0.405\textwidth]{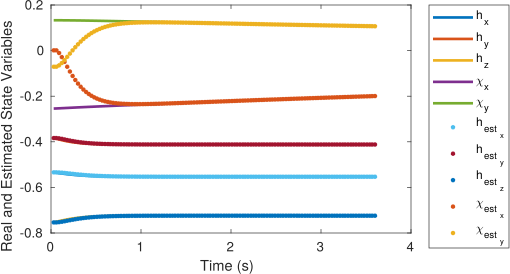}
        \label{fig:sim3a}
    }\qquad
    \subfloat[State estimation error over time.]{
        \includegraphics[width=0.43\textwidth]{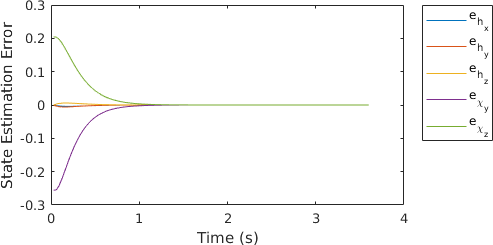}
        \label{fig:sim3b}
    }\\
    \subfloat[Camera's Linear and Angular Velocities.]{
        \includegraphics[width=0.425\textwidth]{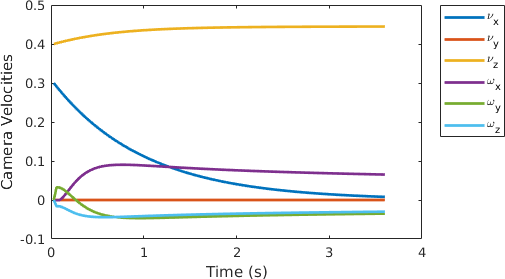}
        \label{fig:sim3c}
    }\qquad
    \subfloat[Evolution of the Eigenvalues of $\bm{\Omega}\bm{\Omega}^T$]{
        \includegraphics[width=0.425\textwidth]{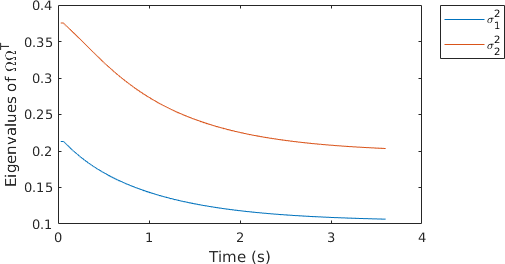}
        \label{fig:sim3d}
    }
    \caption{Simulation results for a single line. The real and the estimated state are presented in the top left plot, and the corresponding state estimation error in the top right plot. The velocities of the camera throughout the task are presented in the bottom left plot. Finally the eigenvalues of the matrix $\bm{\Omega}\bm{\Omega}^T$ are presented in the bottom right plot.}
    \label{fig:sim3}
\end{figure*}

Let us consider the case where the $z$ coordinate of $\bm{\chi}$ is fixed.
Solving the orthogonality constraint for $\chi_z$, we obtain
\begin{equation}
    \chi_z = -\frac{\chi_x h_x + \chi_y h_y}{h_z}.
    \label{eq:chiz}
\end{equation}
By replacing \eqref{eq:chiz} in \eqref{eq:hdyn} and \eqref{eq:chidyn}, and changing the state to $\mathbf{x} = [\mathbf{h},\chi_x,\chi_y]^T$, the dynamics become
\begin{align}
    \dot{\mathbf{h}} = & \bm{\omega}_c \times \mathbf{h} + \bm{\nu}_c^T \mathbf{h} 
    \begin{bmatrix}
        -\frac{h_x h_y}{h_z} & -h_z - \frac{h_y^2}{h_z}\\
        h_z + \frac{h_x^2}{h_z} & \frac{h_x h_y}{h_z}\\
        -h_y & h_x
    \end{bmatrix} \begin{bmatrix} \chi_x \\ \chi_y  \end{bmatrix}
\end{align}

\begin{align}
        \dot{\chi_x} = & -\omega_{c_z} \chi_y - \omega_{c_y} \frac{h_x \chi_x + h_y \chi_y}{h_z} -... \nonumber  \\
        & - \nu_{c_x}\left(h_z \chi_x \chi_y + \frac{ h_y \chi_x (h_x \chi_x + h_y \chi_y)}{h_z}\right) + ... \nonumber \\
        & + \nu_{c_y}\left( h_z \chi_x^2 + \frac{h_x \chi_x (h_x \chi_x + h_y \chi_y)}{h_z}\right) - ... \nonumber \\
        & - \nu_{c_z}(h_y \chi_x^2 - h_x \chi_y \chi_x) \\
        \dot{\chi_y} = & \omega_{c_z} \chi_x + \omega_{c_x} \frac{\chi_x h_x + \chi_y h_y }{h_z} - ... \nonumber \\
        & - \nu_{c_x}\left(h_z \chi_y^2 + \frac{h_y \chi_y (h_x \chi_x + h_y \chi_y)}{h_z}\right) + ... \nonumber \\
        & + \nu_{c_y}\left(h_z \chi_x \chi_y + \frac{(h_x \chi_y (h_x \chi_x + h_y \chi_y)}{h_z}\right) + ... \nonumber \\
        & + \nu_{c_z}(h_y \chi_x^2 - h_y \chi_y \chi_x).
\end{align}
For simplicity sake, let us define $\dot{\chi_x} = f_{u_x}(\mathbf{h},\chi_x,\chi_y,\mathbf{v}_c)$, and $\dot{\chi_y} = f_{u_y}(\mathbf{h},\chi_x,\chi_y,\mathbf{v}_c)$.
These definitions will be used henceforward to refer the dynamics of the unknown variables.
In this formulation 
\begin{equation}
    \bm{\Omega} = \bm{\nu}_c^T \mathbf{h} 
        \begin{bmatrix}
            -\frac{h_x h_y}{h_z} & h_z + \frac{h_x^2}{h_z} & -h_y\\
            -h_z - \frac{h_y^2}{h_z} & \frac{h_x h_y}{h_z} & h_x
        \end{bmatrix},
    \label{eq:omega}
\end{equation}
and $\bm{\Omega}\bm{\Omega}^T$ is full rank, as long as the linear velocity and the moment vector are not orthogonal (this restriction is included in \eqref{eq:Omega_first} also). 
The observer can thus be written as
\begin{equation}
    \begin{cases} 
        \dot{\hat{\mathbf{h}}} = \bm{\omega}_c \times \mathbf{h} + \bm{\Omega}^T \begin{bmatrix} \hat{\chi_x} \\ \hat{\chi_y} \end{bmatrix} + \mathbf{H}\mathbf{\tilde{h}} \\
        \dot{\begin{bmatrix} \hat{\chi_x} \\ \hat{\chi_y} \end{bmatrix}} = \begin{bmatrix}  f_{u_x}(\mathbf{h},\hat{\chi_x},\hat{\chi_y},\mathbf{v}_c) \\ f_{u_y}(\mathbf{h},\hat{\chi_x},\hat{\chi_y},\mathbf{v}_c) \end{bmatrix} + \alpha \bm{\Omega} \mathbf{\tilde{h}}.
    \end{cases}
    \label{eq:obs_z}
\end{equation}
Notice that, the same approach can be replicated for both the $x$ and $y$ coordinates of $\mathbf{h}$, yielding matrices $\bm{\Omega}$ compliant with the requirements for convergence. 
The observer is validated with simulation tests, which will be presented next.

\section{RESULTS}
\label{sec:results}

We start by validating our approach, resorting to simulation tests for a single line.
Then, we apply the method in a real mobile robot \cite{messias2014}, considering a single line, and finally three lines simultaneously. 

\subsection{Simulation Results}
\label{sec:simulation}

This section presents the simulation results for active observation of vector $\bm{\chi}$.
The goal here is to keep the norm of the linear velocity relatively low, so the control effort does not increase excessively, while maintaining the possibility to lead the eigenvalues of $\bm{\Omega}\bm{\Omega}^T$ to a desired value ($\bm{\sigma}_{des}^2$).
Let $\mathbf{J}_{\nu_c} = [\mathbf{J}_{\bm{\nu}_c,1}^T,\mathbf{J}_{\bm{\nu}_c,2}^T]^T \in \mathbb{R}^{2 \times 3}$ be the Jacobian matrix, which relates $\dot{\bm{\nu}}_c$ with the time derivative of the eigenvalues of $\bm{\Omega}\bm{\Omega}^T$, and $\bm{\sigma}^2 = [\sigma_1^2,\sigma_2^2]^T \in \mathbb{R}^2$ be the current eigenvalues. Then, the control law used is 
\begin{equation}
    \dot{\bm{\nu}_c} = k_1 \mathbf{J}_{\bm{\nu}_c}^{\dagger} (\bm{\sigma}_{des}^2 - \bm{\sigma}^2) + k_2 \left( \mathbf{I}_2 - \mathbf{J}_{\bm{\nu}_c}^{\dagger}\mathbf{J}_{\bm{\nu}_c} \right)\bm{\nu}_c,
    \label{eq:control_law}
\end{equation}
 which as been proposed in \cite{spica2013}, where $\mathbf{I}_2 \in \mathbb{R}^{2\times2}$ is an identity matrix, $k_1 > 0$ and $k_2 > 0$ are positive constants, and $\mathbf{J}_{\bm{\nu}_c}^{\dagger}$ is the Moore-Penrose pseudo-inverse of the Jacobian matrix.

In order to compensate the effect of $\dot{\mathbf{h}}$ in the dynamics of the eigenvalues, the condition $\dot{\mathbf{h}} \simeq \mathbf{0}$ is enforced.
This is achieved using the angular velocity, which is obtained by setting \eqref{eq:hdyn} to zero and solving for $\bm{\omega}_c$ yielding
\begin{equation}
    \bm{\omega}_c = (\bm{\nu}_c^T\mathbf{h}) \bm{\hat{\chi}}.
\end{equation}
%Thus the moment vector constant is kept constant, if the camera rotates about the direction of the line with an angle given by $\frac{\bm{\nu}_c^T\mathbf{h}}{d}$. 

The observer in \eqref{eq:obs_z} was simulated in MATLAB, considering a perspective camera, with the intrinsic parameters matrix given by $\mathbf{I}_3 \in \mathbb{R}^{3\times3}$. All six degrees-of-freedom (dof) are assumed to be controllable. 
The true coordinates of the line are obtained by generating a random point in a cube with $3m$ side in front of the camera, then a unit direction is randomly selected, and the moment vector is computed using \eqref{eq:depth_moment}.
Finally, the change of variables (presented in Sec.~\ref{sec:sfm_lines}) is applied.
The initial estimate of ($\hat{\bm{\chi}}$) is also generated randomly, $\hat{\mathbf{h}}$ is initialized with its true value, since it is available from the measurements.
Fig.~\ref{fig:sim3} presents the simulation results for this case, with the following gains $k_1 , k_2 = 1$, $\alpha = 2000$, and $\bm{\sigma}_{des}^2 = [0.1,0.2]^T$ (see \eqref{eq:control_law} and \eqref{eq:obs_z}). 
Fig.~\ref{fig:sim3}\subref{fig:sim3a} presents the state's evolution for both the real system and its estimate, where we can observe that the objective $\dot{\mathbf{h}} \simeq 0$ was achieved.
Fig.~\ref{fig:sim3}\subref{fig:sim3b} presents the state error over time, where we can see that convergence is achieved in about 1 second.
Fig.~\ref{fig:sim3}\subref{fig:sim3c} presents the velocities. Notice that the velocity norm is kept constant.
Finally, Fig.~\ref{fig:sim3}\subref{fig:sim3d} presents the eigenvalues over time. Notice that they reach their desired values.
The total error of the \emph{Pl\"ucker} coordinates is $||\mathbfcal{L} - \mathbfcal{L}_{est}|| = 0.0019$, where  $\mathbfcal{L}$ and $\mathbfcal{L}_{est}$, are the real coordinates, and the estimated coordinates of the line as define in \eqref{eq:biline} respectively.

\subsection{Real Experiments}
\label{sec:real}

Finally, the experimental results with a real robotic platform \cite{messias2014} (see Fig.~\ref{fig:exp_setup}\subref{fig:mbot}) are presented.
It is an omnidirectional mobile platform having 3 dof (2 linear, and 1 angular).
Our observer scheme was implemented resorting to Robot Operating System (\emph{ROS}) \cite{quigley2009}  for sending controls to the platform, and receiving its odometry readings. 
Lines are tracked with the moving-edges tracker \cite{marchand2005b}, available in \emph{ViSP} \cite{marchand2005}. 
The camera used was a Pointgrey Flea3 USB3 \cite{webflea3} mounted on top of the robot, as shown in Fig.~\ref{fig:exp_setup}\subref{fig:mbot}.

\subsubsection{Single line estimation}
\label{sec:single_line}

 \begin{figure}
     \vspace{-.15cm}
     \centering
     \subfloat[Robotic Platform.]{
         \includegraphics[height=0.182\textheight]{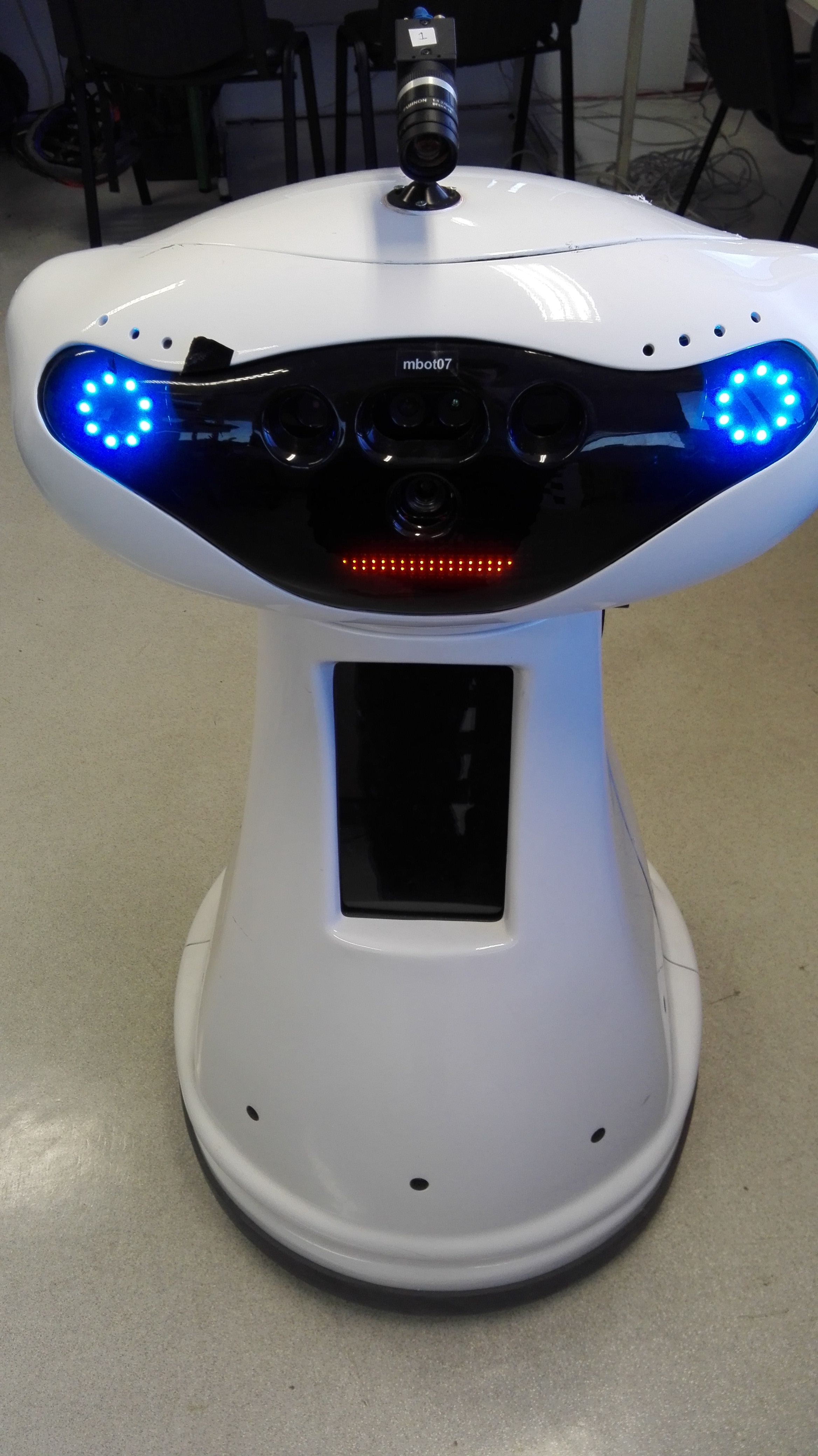}
         \label{fig:mbot}
     }\hfill
     \subfloat[Camera Image with tracked lines and points for pose estimation.]{
         \includegraphics[height=0.182\textheight]{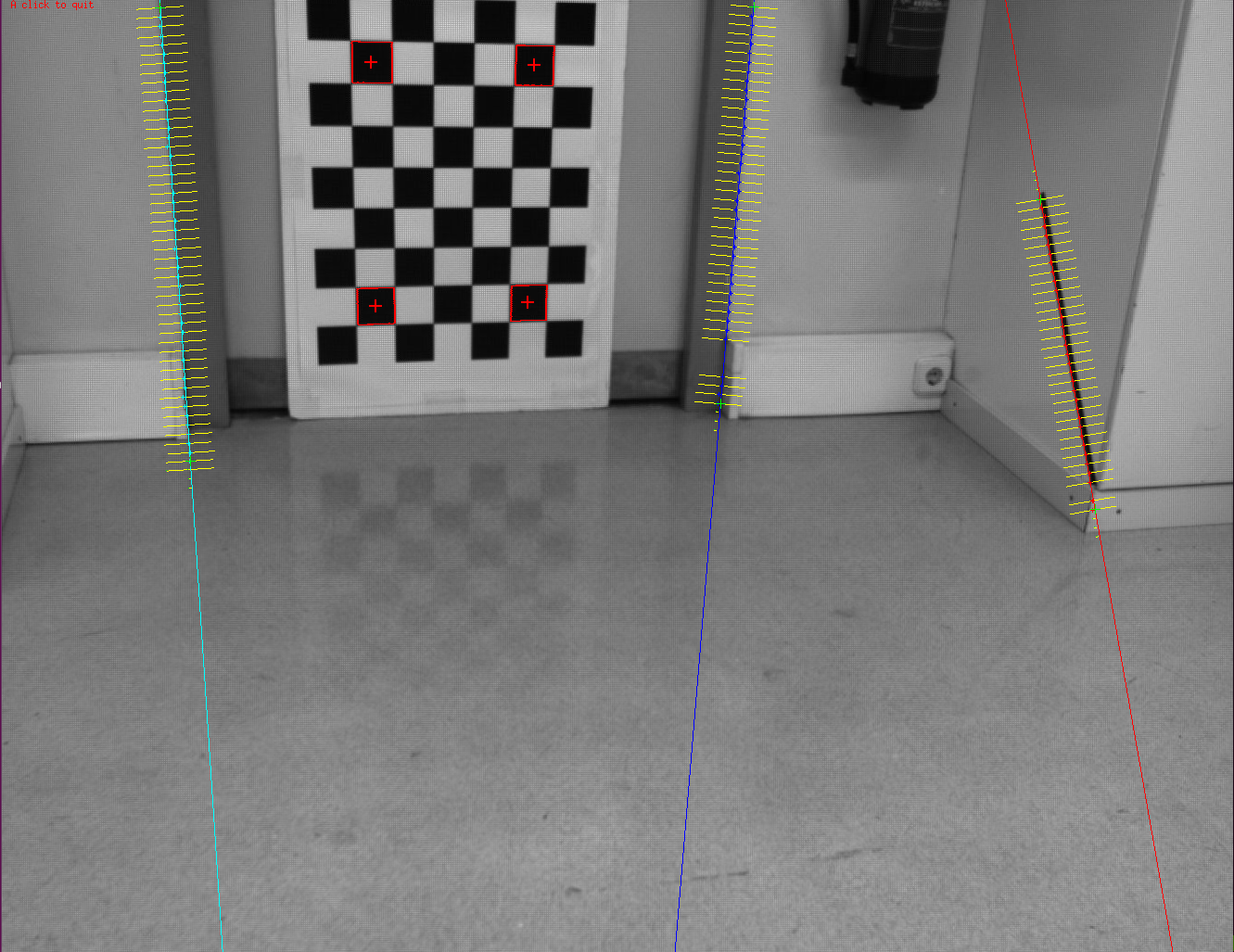}
         \label{fig:line_track}
     }%
     \caption{On the left the robotic platform used in the experimental results is shown. It is an omnidirectional platform, and thus with 3 dof (2 linear, and 1 angular). On top there is a Pointgrey Flea3 USB3 \cite{webflea3} camera, which is the visual sensor used in the experiments. On the right an image from the camera with the four points, of the chessboard, used for pose estimation (to obtain the true coordinates of the lines) and the lines tracked with the moving-edges tracker \cite{marchand2005b}.}
     \label{fig:exp_setup}
 \end{figure}

\begin{figure*}
    \centering
    \subfloat[Estimated state evolution.]{
        \includegraphics[width=0.425\textwidth]{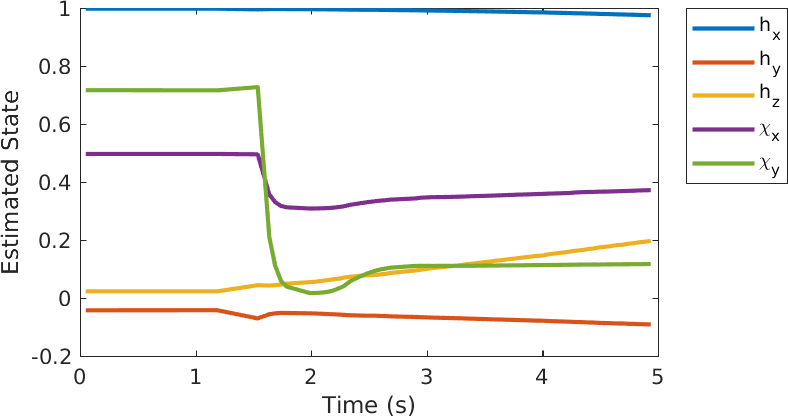}
        \label{fig:real_linea}
    }\qquad
    \subfloat[State estimation error over time.]{
        \includegraphics[width=0.425\textwidth]{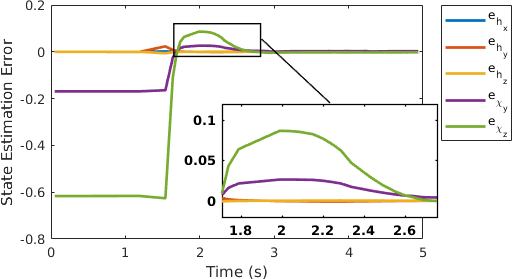}
        \label{fig:real_lineb}
    }\\
    \subfloat[Camera's Linear and Angular Velocities.]{
        \includegraphics[width=0.425\textwidth]{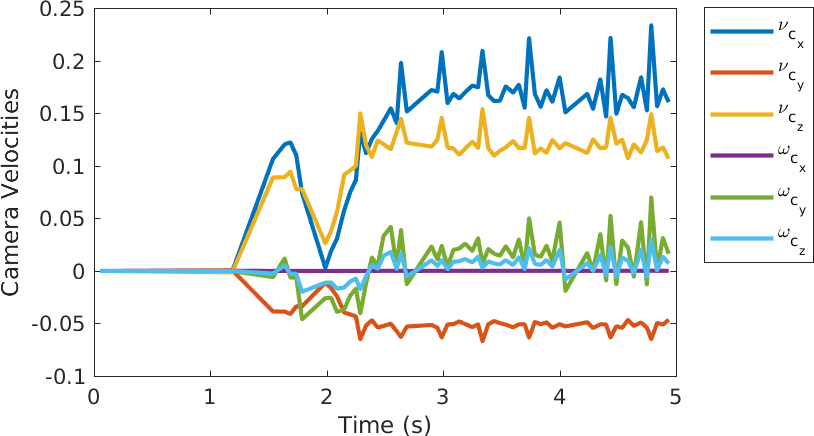}
        \label{fig:real_linec}
    }\qquad
    \subfloat[Evolution of the Eigenvalues of the matrix $\bm{\Omega}\bm{\Omega}^T$]{
        \includegraphics[width=0.425\textwidth]{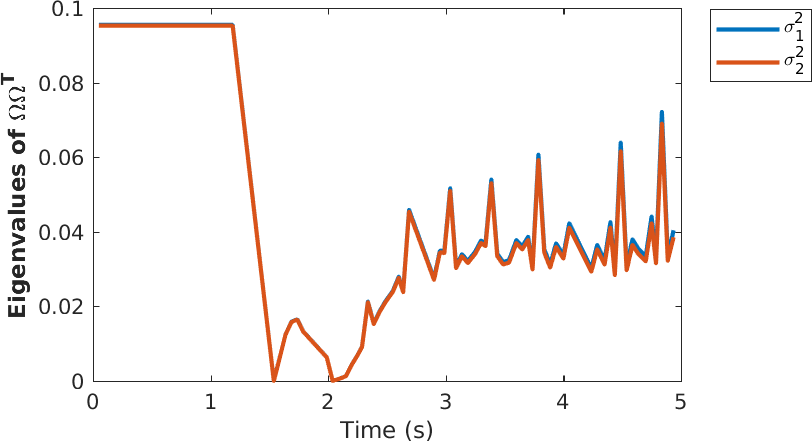}
        \label{fig:real_lined}
    }%
    \caption{Experimental results with a real robot and camera for a single line. The estimated state is presented in the top left plot, and the corresponding state estimation error in the top right plot. The velocities of the camera throughout the task are presented in the bottom left plot. Finally, the eigenvalues of the matrix $\bm{\Omega}\bm{\Omega}^T$ are presented in the bottom right plot.}
    \label{fig:real_line}
\end{figure*}

In a first step, we replicate the experimental setup in Sec.~\ref{sec:simulation}, in order to show that the performance is kept in a real setup.
The goal is to estimate a line defined in a chessboard. 
The board is used only to obtain the initial coordinates of the line, by computing the initial pose of the camera w.r.t. the board. 
For this purpose, we used four points (marked in the Fig.~\ref{fig:exp_setup}\subref{fig:line_track}), and then applied the \emph{POSIT} algorithm \cite{oberkampf1996} available in \emph{ViSP}.
The camera's velocity is retrieved from the robot's odometry, after applying the transformation from the robot's base to the camera reference frame (computed a priori). Fig.~\ref{fig:real_line} presents the experimental results for this case, with the following gains $k_1 , k_2 = 1$, $\alpha = 2000$, and $\bm{\sigma}_{des}^2 = [0.1,0.2]^T$.
Fig.~\ref{fig:real_line}\subref{fig:real_linea} presents the estimated state evolution, where we can observe that the objective $\dot{\mathbf{h}} \simeq \mathbf{0}$ is not entirely achieved. This is due to the reduced number of dof.
Fig.~\ref{fig:real_line}\subref{fig:real_lineb} presents the state error over time. We can see that convergence is achieved in about 3 seconds. Notice that it takes some time for the robot to start moving, this is a limitation on the robot's drivers.
Fig.~\ref{fig:real_line}\subref{fig:real_linec} presents the velocities, where we note that the velocity readings are noisy, since they are provided by the robot's odometry.
Finally Fig.~\ref{fig:real_line}\subref{fig:real_lined} presents the eigenvalues, which given the noisy velocity readings did not reached the desired values.
The total error is $||\mathbfcal{L} - \mathbfcal{L}_{est}|| = 0.0067$.

\subsubsection{Estimation of three lines}
\label{sec:3lines}

This section presents the results for three different straight lines.
In this case $m = 9$ and $p = 6$, meaning that we have nine measurements and six unknowns, then $\bm{\Omega}\bm{\Omega}^T \in \mathbb{R}^{6\times6}$ (see the definition of the state in the beginning of Sec.~\ref{sec:active}).
We considered the variables to optimize to be the mean of the eigenvalues associated with each line, thus
\begin{equation}
    \bm{\sigma}_{3 \, lines}^2 = \frac{1}{3} \sum_{i=1}^{3} \bm{\sigma}_{i}^2,
\end{equation}
where $\bm{\sigma}_{i}^2 = \left[\sigma_{i,1}^2, \sigma_{i,2}^2\right]^T$.
The Jacobian matrix, which yields the relation between the new eigenvalues and the camera's linear velocity is then
\begin{equation}
    \mathbf{J}_{\bm{\nu}_c,3 \, lines} = \frac{1}{3} \sum_{i = 1}^{3} \mathbf{J}_{\bm{\nu}_c,i}.
\end{equation}

\begin{figure*}
    %\vspace{-.15cm}
    \centering
    \subfloat[State Estimation Error First Line, which is the red line in the bottom plot.]{
        \includegraphics[width=0.3\textwidth]{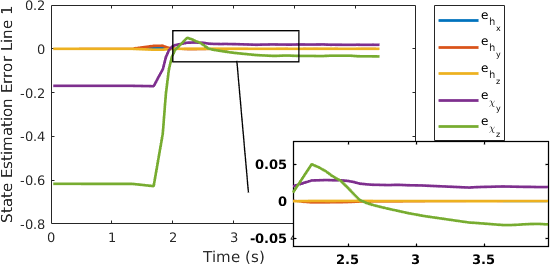}
        \label{fig:real_multi_linea}
    } \,\,
    \subfloat[State Estimation Error Second Line, which is the cyan line in the bottom plot.]{
        \includegraphics[width=0.3\textwidth]{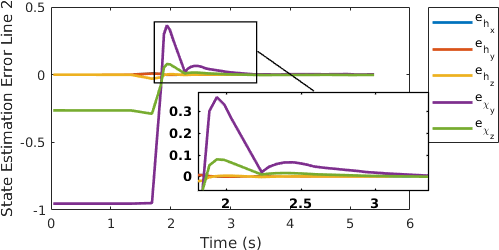}
        \label{fig:real_multi_lineb}
    } \,\,
    \subfloat[State Estimation Error Third Line, which is the blue line in the bottom plot.]{
        \includegraphics[width=0.3\textwidth]{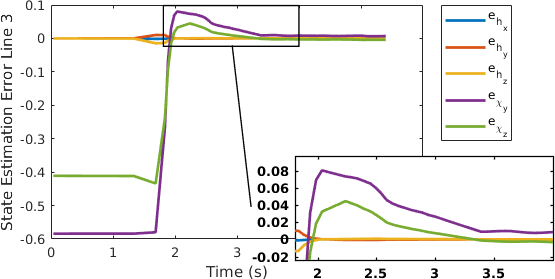}
        \label{fig:real_multi_linec}
    }
    \newline
    \subfloat[Camera Velocities.]{
        \includegraphics[width=0.45\textwidth]{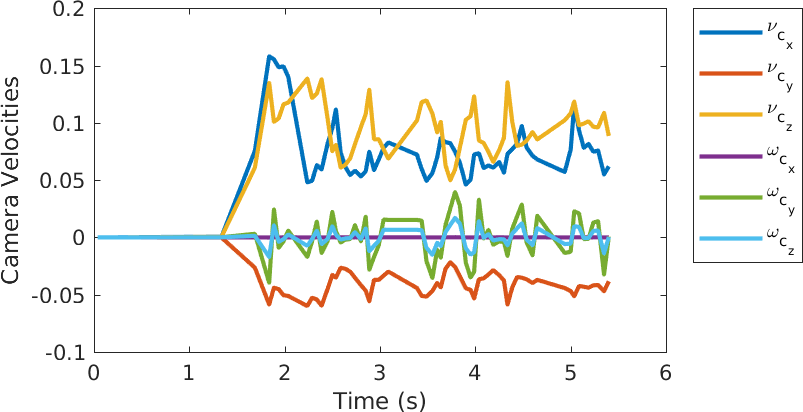}
        \label{fig:real_multi_lined}
    } \qquad
    \subfloat[Estimated  and real lines in the robot's reference frame. Besides it is shown the camera's final position and its path.]{
        \includegraphics[width=0.37\textwidth]{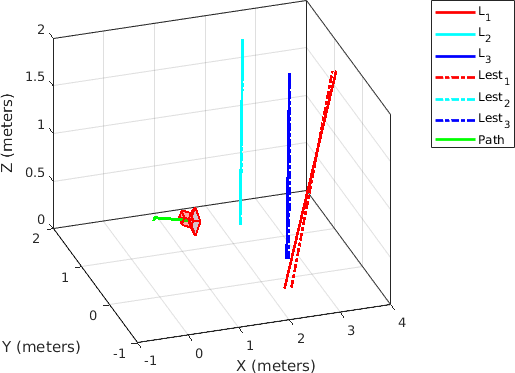}
        \label{fig:real_multi_linee}
    }
    %\subfloat[Robotic Platform.]{
    %    \includegraphics[height=0.17\textheight]{mbot.jpg}
    %    \label{fig:mbot}
    %}\hfill
    %\subfloat[Camera Image with tracked lines and points for pose estimation.]{
    %    \includegraphics[height=0.17\textheight]{cam_image.png}
    %    \label{fig:line_track}
    %}\,
    \caption{Experimental results with a real robot and camera for three lines. The estimation state errors are presented for each line (top three plots). The velocities of the camera throughout the task are presented in the bottom left plot. A 3D plot of the estimated and real lines, alongside the camera and its path is presented in the bottom right.}
    \label{fig:real_multi_line}
\end{figure*}

The experimental setup is similar to the one in Sec.~\ref{sec:single_line}.
A chessboard is used to obtain the initial \emph{Pl\"ucker} coordinates of the 3D straight lines.
Throughout the experiment lines are tracked with the moving edges tracker of \emph{ViSP}, the tracking of the lines in a camera picture is presented in Fig.~\ref{fig:exp_setup}\subref{fig:line_track}. Fig.~\ref{fig:real_multi_line} presents the experimental results for this case, with the following gains $k_1 , k_2 = 1$, $\alpha = 1000$, and $\bm{\sigma}_{des}^2 = [0.1,0.2]^T$.
Figs.~\ref{fig:real_multi_line}\subref{fig:real_multi_linea}, ~\ref{fig:real_multi_line}\subref{fig:real_multi_lineb}, and ~\ref{fig:real_multi_line}\subref{fig:real_multi_linec} present the state error estimation for the first, second, and third line respectively. 
Fig.~\ref{fig:real_multi_line}\subref{fig:real_multi_lined} presents the camera's velocities, which are noisy, since they are retrieved from the robot's odometry. 
Finally, Fig.~\ref{fig:real_multi_line}\subref{fig:real_multi_linee} presents a 3D plot with the position of each line (real and estimated coordinates). Besides it presents the final position of the camera, and its path.
The total error of the \emph{Pl\"ucker} coordinates are $||\mathbfcal{L}_1 - \mathbfcal{L}_{est,1}|| = 0.0388$, $||\mathbfcal{L}_1 - \mathbfcal{L}_{est,1}|| = 0.0109$, and $||\mathbfcal{L}_1 - \mathbfcal{L}_{est,1}|| = 0.0217$.

\section{CONCLUSIONS}
\label{sec:conclusions}

This paper has proposed a strategy for \emph{Active Structure-from-Motion} for 3D straight lines. 
Contrarily to previous approaches, \emph{binormalized Pl\"ucker
coordinates} were used to represent lines, which allow for an explicit representation.
A variable change of these coordinates has been presented to comply with the requirements of a recently presented framework for \emph{Active Structure-from-Motion}.
The new dynamics of the line, and an observer for retrieving the lines' 3D structure were then proposed.
A control law was used with the purpose of keeping the control effort relatively low,  while  achieving  a  desired  convergence  rate.
This approach was validated in simulation and afterwards in a real robotic platform, for both one and three lines.
Future work consists in including both the algebraic constraints on the \emph{Pl\"ucker} coordinates in the estimation scheme, and the observer in a Image-Based Visual Servoing control scheme.


%%%%%%%%%%%%%%%%%%%%%%%%%%%%%%%%%%%%%%%%%%%%%%%%%%%%%%%%%%%%%%%%%%%%%%%%%%%%%%%%
%\section*{APPENDIX}


%\section*{ACKNOWLEDGMENT}
%
%The preferred spelling of the word �acknowledgment� in America is without an �e� after the �g�. Avoid the stilted expression, �One of us (R. B. G.) thanks . . .�  Instead, try �R. B. G. thanks�. Put sponsor acknowledgments in the unnumbered footnote on the first page.

%%%%%%%%%%%%%%%%%%%%%%%%%%%%%%%%%%%%%%%%%%%%%%%%%%%%%%%%%%%%%%%%%%%%%%%%%%%%%%%%

\bibliographystyle{IEEEtran}
\bibliography{active_sfm}

\end{document}